\documentclass{article}

\PassOptionsToPackage{round}{natbib}
\PassOptionsToPackage{colorlinks,allcolors=black,breaklinks}{hyperref}

\usepackage[final]{neurips_2019}

\usepackage[utf8]{inputenc} % allow utf-8 input
\usepackage[T1]{fontenc}    % use 8-bit T1 fonts
\usepackage{hyperref}       % hyperlinks
\usepackage{url}            % simple URL typesetting
\usepackage{booktabs}       % professional-quality tables
\usepackage{amsfonts}       % blackboard math symbols
\usepackage{nicefrac}       % compact symbols for 1/2, etc.
\usepackage{microtype}      % microtypography
\usepackage[dvipsnames]{xcolor}  % color
\definecolor{mypurple}{RGB}{150, 34, 173}
\definecolor{mygreen}{RGB}{96, 128, 0}
\usepackage{enumitem}       % enumeration
\usepackage{pbox}
\usepackage{graphicx}
\usepackage{subcaption}
\usepackage{multirow}

\title{SuperGLUE: A Stickier Benchmark for General-Purpose Language Understanding Systems}

\author{%
  Alex Wang\thanks{Equal contribution. Correspondence: \texttt{glue-benchmark-admin@googlegroups.com}} \\
  New York University\\
\And
  Yada Pruksachatkun$^*$ \\
  New York University\\
\And
  Nikita Nangia$^*$ \\
  New York University\\
\And
  Amanpreet Singh$^*$ \\
  Facebook AI Research\\
\And
  Julian Michael \\
  University of Washington\\
\And
  Felix Hill \\ 
  DeepMind\\
\And
  Omer Levy \\
  Facebook AI Research\\
\And
  Samuel R. Bowman \\
  New York University \\
}

\begin{document}

\maketitle

\begin{abstract}

In the last year, new models and methods for pretraining and transfer learning have driven striking performance improvements across a range of language understanding tasks. The GLUE benchmark, introduced a little over one year ago, offers a single-number metric that summarizes progress on a diverse set of such tasks, but performance on the benchmark has recently surpassed the level of non-expert humans, suggesting limited headroom for further research. In this paper we present SuperGLUE, a new benchmark styled after GLUE with a new set of more difficult language understanding tasks, a software toolkit, and a public leaderboard.
SuperGLUE is available at \href{https://super.gluebenchmark.com/}{\tt super.gluebenchmark.com}.

\end{abstract}

\section{Introduction}

Recently there has been notable progress across many natural language processing (NLP) tasks, led by methods such as ELMo \citep{peters2018deep}, OpenAI GPT \citep{radford2018improving}, and BERT \citep{devlin2018bert}. The unifying theme of these methods is that they couple self-supervised learning from massive unlabelled text corpora with effective adapting of the resulting model to target tasks. The tasks that have proven amenable to this general approach include question answering, textual entailment, and parsing, among many others \citep[][i.a.]{devlin2018bert,kitaev2018multilingual}. 
%Besides their striking gains in performance on many such tasks, both ELMo and BERT have been recognized with best paper awards at major conferences and widespread deployment in industry.

In this context, the GLUE benchmark \citep{wang2018glue} has become a prominent evaluation framework for research towards general-purpose language understanding technologies. 
GLUE is a collection of nine language understanding tasks built on existing public datasets, together with private test data, an evaluation server, a single-number target metric, and an accompanying expert-constructed diagnostic set. GLUE was designed to provide a general-purpose evaluation of language understanding that covers a range of training data volumes, task genres, and task formulations. We believe it was these aspects that made GLUE particularly appropriate for exhibiting the transfer-learning potential of approaches like OpenAI GPT and BERT.

The progress of the last twelve months has eroded headroom on the GLUE benchmark dramatically. 
While some tasks (Figure~\ref{fig:benchmark-trends}) and some linguistic phenomena (Figure~\ref{fig:diagnostic-trends} in Appendix~\ref{ax:diagnostics}) measured in GLUE remain difficult, 
%the current state of the art GLUE Score \citep[83.8 with the BERT-based MT-DNN system from][]{liu2019mt} is only 3.3 points behind human performance \citep[87.1 from][]{nangia2019human}, and in fact exceeds this human performance estimate on three tasks.
the current state of the art GLUE Score as of early July 2019
%\citep[88.4 with the XLNet system from][]{yang2019xlnet} 
\citep[88.4 from][]{yang2019xlnet}
surpasses human performance \citep[87.1 from][]{nangia2019human} by 1.3 points, and in fact exceeds this human performance estimate on four tasks.
%\footnote{The Quora Question Pairs, The Microsoft Research Paraphrase Corpus \citep{dolan2005automatically}, and QNLI, an answer sentence selection task derived from SQuAD \citep{rajpurkar2016}.} 
Consequently, while there remains substantial scope for improvement towards GLUE's high-level goals, the original version of the benchmark is no longer a suitable metric for quantifying such progress.

\begin{figure}
    \centering
    \includegraphics[width=0.97\textwidth]{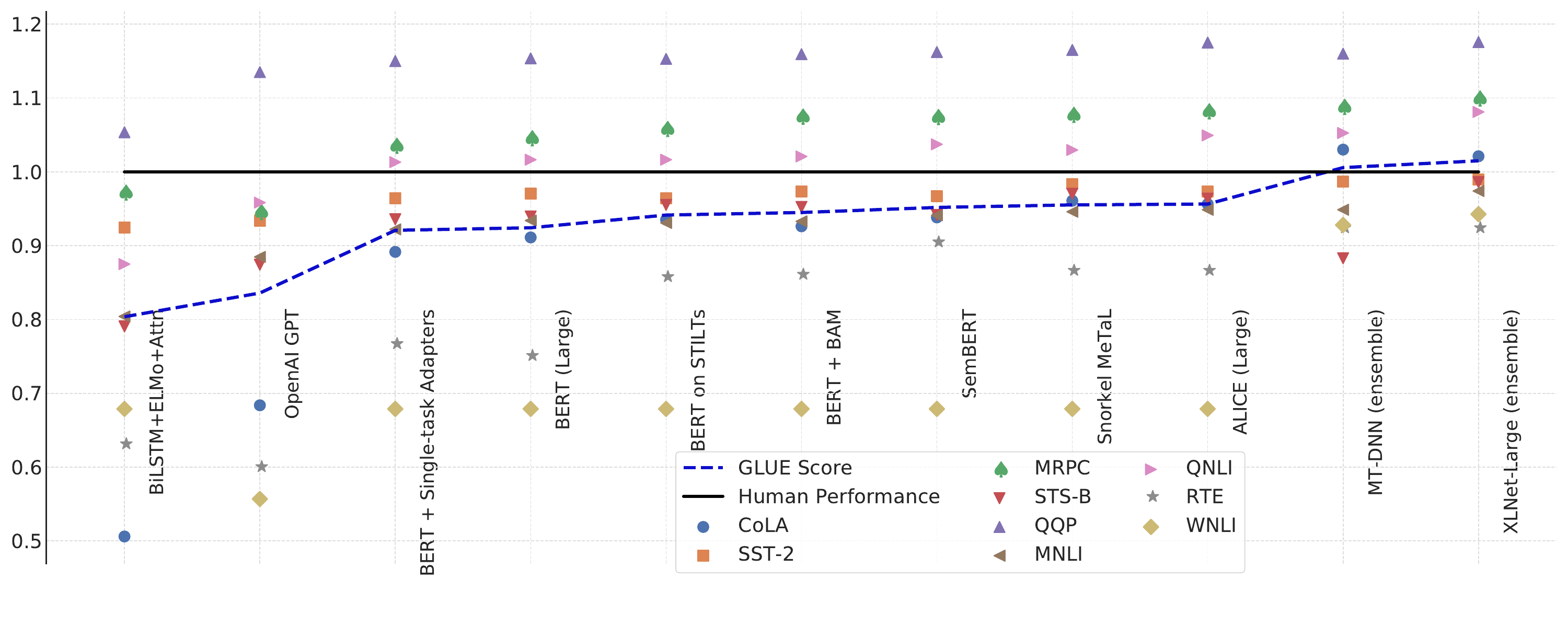}
    \caption{GLUE benchmark performance for submitted systems, rescaled to set human performance to 1.0, shown as a single number score, and broken down into the nine constituent task performances. For tasks with multiple metrics, we use an average of the metrics. More information on the tasks included in GLUE can be found in \citet{wang2018glue} and in \citet[CoLA]{warstadt2018neural}, \citet[SST-2]{socher2013recursive}, \citet[MRPC]{dolan2005automatically}, \citet[STS-B]{cer-etal-2017-semeval}, and \citet[MNLI]{williams2018broad}, and \citet[the original data source for QNLI]{rajpurkar2016}.}
    \label{fig:benchmark-trends}
\end{figure}

In response, we introduce SuperGLUE, a new benchmark designed to pose a more rigorous test of language understanding. SuperGLUE has the same high-level motivation as GLUE: to provide a simple, hard-to-game measure of progress toward general-purpose language understanding technologies for English. We anticipate that significant progress on SuperGLUE should require substantive innovations in a number of core areas of machine learning, including sample-efficient, transfer, multitask, and unsupervised or self-supervised learning.
%We believe SuperGLUE is of interest to the machine learning community as a robust testbed for validating task-agnostic improvements in these areas on a difficult open problem.

SuperGLUE follows the basic design of GLUE: It consists of a public leaderboard built around eight language understanding tasks, drawing on existing data, accompanied by a single-number performance metric, and an analysis toolkit. 
However, it improves upon GLUE in several ways:

%\begin{itemize}[]
%    \item \textbf{More challenging tasks:} SuperGLUE retains the two hardest tasks in GLUE. The remaining tasks were identified from those submitted to an open call for task proposals and were selected based on difficulty for current NLP approaches.%\footnote{\href{http://bit.ly/glue2cfp}{\tt bit.ly/glue2cfp}} 
%    \item\textbf{More diverse task formats:} The task formats in GLUE are limited to sentence- and sentence-pair classification. We expand the set of task formats in SuperGLUE to include coreference resolution and question answering (QA). 
%    \item \textbf{Comprehensive human baselines:} We include human performance estimates for all benchmark tasks, which verify that substantial headroom exists between a strong BERT-based baseline and human performance.
%    \item \textbf{Improved code support:} SuperGLUE is distributed with a new, modular toolkit for work on pretraining, multi-task learning, and transfer learning in NLP, built around standard tools including PyTorch \citep{paszke2017automatic} and AllenNLP \citep{Gardner2017AllenNLP}.
%    \item \textbf{Refined usage rules:} The conditions for inclusion on the SuperGLUE leaderboard 
%    %differ from those governing GLUE in several ways, all meant 
%    have been revamped to ensure fair competition, an informative leaderboard, and full credit assignment to data and task creators.
%\end{itemize}

\textbf{More challenging tasks:} SuperGLUE retains the two hardest tasks in GLUE. The remaining tasks were identified from those submitted to an open call for task proposals and were selected based on difficulty for current NLP approaches.%\footnote{\href{http://bit.ly/glue2cfp}{\tt bit.ly/glue2cfp}} 

\textbf{More diverse task formats:} The task formats in GLUE are limited to sentence- and sentence-pair classification. We expand the set of task formats in SuperGLUE to include coreference resolution and question answering (QA). 

\textbf{Comprehensive human baselines:} We include human performance estimates for all benchmark tasks, which verify that substantial headroom exists between a strong BERT-based baseline and human performance.

\textbf{Improved code support:} SuperGLUE is distributed with a new, modular toolkit for work on pretraining, multi-task learning, and transfer learning in NLP, built around standard tools including PyTorch \citep{paszke2017automatic} and AllenNLP \citep{Gardner2017AllenNLP}.

\textbf{Refined usage rules:} The conditions for inclusion on the SuperGLUE leaderboard have been revamped to ensure fair competition, an informative leaderboard, and full credit assignment to data and task creators.

The SuperGLUE leaderboard, data, and software tools are available at \href{https://super.gluebenchmark.com/}{\tt super.gluebenchmark.com}.

\section{Related Work}

Much work prior to GLUE demonstrated that training neural models with large amounts of available supervision can produce representations that effectively transfer to a broad range of NLP tasks \citep{collobert2008unified,dai2015semisupervised,kiros2015skip,hill2016learning,conneau2018senteval,mccann2017learned,peters2018deep}.
GLUE was presented as a formal challenge affording straightforward comparison between such task-agnostic transfer learning techniques.
Other similarly-motivated benchmarks include SentEval~\citep{conneau2018senteval}, which specifically evaluates fixed-size sentence embeddings, and DecaNLP~\citep{mccann2018decanlp}, which recasts a set of target tasks into a general question-answering format and prohibits task-specific parameters. In contrast, GLUE provides a lightweight classification API and no restrictions on model architecture or parameter sharing, which seems to have been well-suited to recent work in this area.

Since its release, GLUE has been used as a testbed and showcase by the developers of several influential models, including GPT~\citep{radford2018improving} and BERT~\citep{devlin2018bert}. As shown in Figure~\ref{fig:benchmark-trends}, progress on GLUE since its release has been striking.
On GLUE, GPT and BERT achieved scores of 72.8 and 80.2 respectively, relative to 66.5 for an ELMo-based model \citep{peters2018deep} and 63.7 for the strongest baseline with no multitask learning or pretraining above the word level. 
%With further progress stimulated by even larger models \citep{yang2019xlnet}, 
Recent models \citep{liu2019mt,yang2019xlnet} have clearly surpassed estimates of non-expert human performance on GLUE \citep{nangia2019human}. 
The success of these models on GLUE has been driven by ever-increasing model capacity, compute power, and data quantity, as well as innovations in model expressivity (from recurrent to bidirectional recurrent to multi-headed transformer encoders) and degree of contextualization (from learning representation of words in isolation to using uni-directional contexts and ultimately to leveraging bidirectional contexts).
In parallel to work scaling up pretrained models, several studies have focused on complementary methods for augmenting performance of pretrained models. \citet{phang2018sentence} show that BERT can be improved using two-stage pretraining, i.e., fine-tuning the pretrained model on an intermediate data-rich supervised task before fine-tuning it again on a data-poor target task.
\citet{liu2019mt, liu2019improving} and \citet{snorkel:2018} get further improvements respectively via multi-task finetuning and using massive amounts of weak supervision.
\citet{clark2019bam} demonstrate that knowledge distillation \citep{hinton2015distilling,furlanello2018born} can lead to student networks that outperform their teachers.
%, leading to further improvements over BERT.
Overall, the quantity and quality of research contributions aimed at the challenges posed by GLUE underline the utility of this style of benchmark for machine learning researchers looking to evaluate new application-agnostic methods on language understanding.

% \begin{figure}\centering
%     \includegraphics[width=0.8\textwidth]{images/diagnostics.pdf}
%     % diagnostic trends plot
%     \caption{Performance of GLUE submissions on selected diagnostic categories, reported using the $R_3$ metric scaled up by 100, as in \citet[see paper for a description of the categories]{wang2018glue}.
%     While some initially difficult categories saw gains from advances on GLUE (e.g., double negation), others remained hard (restrictivity) or even adversarial (disjunction, downward monotone).}
%     \label{fig:diagnostic-trends}
% \end{figure}

Limits to current approaches are also apparent via the GLUE suite. 
%Top performance on Winograd-NLI \citep[based on][]{levesque2012winograd} is still at the majority baseline, with accuracy (65.1) far below human-level (95.9). 
Performance on the GLUE diagnostic entailment dataset, at 0.42 $R_3$, falls far below the average human performance of 0.80 $R_3$ reported in the original GLUE publication, with models performing near, or even below, chance on some linguistic phenomena (Figure~\ref{fig:diagnostic-trends}, Appendix~\ref{ax:diagnostics}). While some initially difficult categories saw gains from advances on GLUE (e.g., double negation), others remain hard (restrictivity) or even adversarial (disjunction, downward monotonicity). This suggests that even as unsupervised pretraining produces ever-better statistical summaries of text, it remains difficult to extract many details crucial to semantics without the right kind of supervision. 
Much recent work has made similar observations about the limitations of existing pretrained models
\citep{jia2017adversarial, naik2018stresstest, mccoy2019nonentailed, mccoy2019wrongreasons,  liu2019transferability, liu2019inoculation}.

\section{SuperGLUE Overview}

\begin{table*}[t]
\caption{The tasks included in SuperGLUE.  
\textit{WSD} stands for word sense disambiguation, \textit{NLI} is natural language inference, \textit{coref.} is coreference resolution, and \textit{QA} is question answering. For MultiRC, we list the number of total answers for 456/83/166 train/dev/test questions. 
%The metrics for MultiRC are binary F1 on all answer-options and exact match.
}
\centering \small
\begin{tabular}{lrrrlll}
 \toprule
\textbf{Corpus} & \textbf{$|$Train$|$} & \textbf{$|$Dev$|$} & \textbf{$|$Test$|$} & \textbf{Task} & \textbf{Metrics} & \textbf{Text Sources} \\
\midrule 
BoolQ & 9427 & 3270 & 3245 & QA & acc. & Google queries, Wikipedia \\
CB & 250 & 57 & 250 & NLI & acc./F1 & various \\
COPA & 400 & 100 & 500 & QA & acc. & blogs, photography encyclopedia\\
MultiRC & 5100 & 953 & 1800 & QA & F1$_a$/EM & various \\
ReCoRD & 101k & 10k & 10k & QA & F1/EM & news (CNN, Daily Mail) \\
RTE & 2500 & 278 & 300 & NLI & acc. & news, Wikipedia \\
WiC & 6000 & 638 & 1400 & WSD & acc. & WordNet, VerbNet, Wiktionary \\
WSC & 554 & 104 & 146 & coref. & acc. & fiction books \\
\bottomrule
\end{tabular}
\label{tab:tasks}
\end{table*}

\begin{table*}[t]
\caption{Development set examples from the tasks in SuperGLUE. \textbf{Bold} text represents part of the example format for each task. Text in \textit{italics} is part of the model input. \underline{\textit{Underlined}} text is specially marked in the input. Text in a \texttt{monospaced font} represents the expected model output.}
\label{tab:examples}
\centering \footnotesize
\begin{tabular}{p{0.005\textwidth}p{0.93\textwidth}}

 \toprule
 \parbox[t]{1mm}{\multirow{2}{*}{\rotatebox[origin=c]{90}{{\textbf{BoolQ}}}}} &
\textbf{Passage:} \textit{Barq's -- Barq's is an American soft drink. Its brand of root beer is notable for having caffeine. Barq's, created by Edward Barq and bottled since the turn of the 20th century, is owned by the Barq family but bottled by the Coca-Cola Company. It was known as Barq's Famous Olde Tyme Root Beer until 2012.} \\ & \textbf{Question:} \textit{is barq's root beer a pepsi product} \quad \textbf{Answer:} \texttt{No}\\

\midrule
\parbox[t]{1mm}{\multirow{2}{*}{\rotatebox[origin=c]{90}{{\textbf{CB}}}}} &
\textbf{Text:} \textit{B: And yet, uh, I we-, I hope to see employer based, you know, helping out. You know, child, uh, care centers at the place of employment and things like that, that will help out. A: Uh-huh. B: What do you think, do you think we are, setting a trend?} \\ & \textbf{Hypothesis:} \textit{they are setting a trend} \quad \textbf{Entailment:} \texttt{Unknown}\\

\midrule
\parbox[t]{1mm}{\multirow{2}{*}{\rotatebox[origin=c]{90}{{\textbf{COPA}}}}} & \textbf{Premise:} \textit{My body cast a shadow over the grass.}\quad
\textbf{Question:} \textit{What’s the CAUSE for this?}\\
&\textbf{Alternative 1:} \textit{The sun was rising.}\quad
\textbf{Alternative 2:} \textit{The grass was cut.}\\
&\textbf{Correct Alternative:} \texttt{1}\\

\midrule
\parbox[t]{1mm}{\multirow{2}{*}{\rotatebox[origin=c]{90}{{\textbf{MultiRC}}}}} &
\textbf{Paragraph:} \textit{Susan wanted to have a birthday party. She called all of her friends. She has five friends. Her mom said that Susan can invite them all to the party. Her first friend could not go to the party because she was sick. Her second friend was going out of town. Her third friend was not so sure if her parents would let her. The fourth friend said maybe. The fifth friend could go to the party for sure. Susan was a little sad. On the day of the party, all five friends showed up. Each friend had a present for Susan. Susan was happy and sent each friend a thank you card the next week}\\
& \textbf{Question:} \textit{Did Susan's sick friend recover?} \textbf{Candidate answers:} 
\textit{Yes, she recovered} (\texttt{T}), 
\textit{No} (\texttt{F}), 
\textit{Yes} (\texttt{T}), 
\textit{No, she didn't recover} (\texttt{F}), 
\textit{Yes, she was at Susan's party} (\texttt{T})\\

\midrule
\parbox[t]{1mm}{\multirow{2}{*}{\rotatebox[origin=c]{90}{{\textbf{ReCoRD}}}}} & 
\textbf{Paragraph:} \textit{(\underline{CNN}) \underline{Puerto Rico} on Sunday overwhelmingly voted for statehood. But Congress, the only body that can approve new states, will ultimately decide whether the status of the \underline{US} commonwealth changes. Ninety-seven percent of the votes in the nonbinding referendum favored statehood, an increase over the results of a 2012 referendum, official results from the \underline{State Electorcal Commission} show. It was the fifth such vote on statehood. "Today, we the people of \underline{Puerto Rico} are sending a strong and clear message to the \underline{US Congress}  ... and to the world ... claiming our equal rights as \underline{American} citizens, \underline{Puerto Rico} Gov. \underline{Ricardo Rossello} said in a news release. @highlight \underline{Puerto Rico} voted Sunday in favor of \underline{US} statehood}\\
&\textbf{Query} For one, they can truthfully say, ``Don't blame me, I didn't vote for them, '' when discussing the <placeholder> presidency \quad \textbf{Correct Entities:} \texttt{US} \\

\midrule
\parbox[t]{1mm}{\multirow{2}{*}{\rotatebox[origin=c]{90}{{\textbf{RTE}}}}} &
\textbf{Text:} \textit{Dana Reeve, the widow of the actor Christopher Reeve, has died of lung cancer at age 44, according to the Christopher Reeve Foundation.}\\
& \textbf{Hypothesis:} \textit{Christopher Reeve had an accident.} \quad
\textbf{Entailment:} \texttt{False}\\

\midrule
\parbox[t]{1mm}{\multirow{2}{*}{\rotatebox[origin=c]{90}{{\textbf{WiC}}}}} &
\textbf{Context 1:} \textit{Room and \underline{board}.} \quad
\textbf{Context 2:} \textit{He nailed \underline{boards} across the windows.} \\
& \textbf{Sense match:} \texttt{False}\\

\midrule
\parbox[t]{1mm}{\multirow{1}{*}{\rotatebox[origin=c]{90}{{\textbf{WSC}}}}} & 
\textbf{Text:} \textit{Mark told \underline{Pete} many lies about himself, which Pete included in his book. \underline{He} should have been more truthful.} \quad \textbf{Coreference:} \texttt{False}\vspace{0.25em}\\
\bottomrule
\end{tabular}

\end{table*}

\subsection{Design Process}
%Although models have overtaken estimates of non-expert human performance on GLUE \citep{nangia2019human}, it seems unlikely that a machine capable of robust, human-level language understanding will emerge soon. To create a more challenging and stickier benchmark, we aim to focus SuperGLUE on datasets like Winograd-NLI: language tasks that are simple and intuitive for non-specialist humans but pose a significant challenge to BERT and its friends. To this end, we identify a set of criteria for selecting tasks to include in SuperGLUE
%Although models are fast approaching estimates of non-expert human performance on GLUE \citep{nangia2019human}, 

The goal of SuperGLUE is to provide a simple, robust evaluation metric of any method capable of being applied to a broad range of language understanding tasks. To that end, in designing SuperGLUE, we identify the following desiderata of tasks in the benchmark:

\textbf{Task substance:} Tasks should test a system's ability to understand and reason about texts in English. 

\textbf{Task difficulty:} Tasks should be beyond the scope of current state-of-the-art systems, but solvable by most college-educated English speakers. We exclude tasks that require domain-specific knowledge, e.g. medical notes or scientific papers.

\textbf{Evaluability:} Tasks must have an automatic performance metric that corresponds well to human judgments of output quality. Some text generation tasks fail to meet this criteria due to issues with automatic metrics like ROUGE and BLEU \citep[][i.a.]{callison2006re,liu2016not}.

\textbf{Public data:} We require that tasks have \textit{existing} public training data in order to minimize the risks involved in newly-created datasets. We also prefer tasks for which we have access to (or could create) a test set with private labels.

\textbf{Task format:} We prefer tasks that had relatively simple input and output formats, to avoid incentivizing the users of the benchmark to create complex task-specific model architectures. Still, while GLUE is restricted to tasks involving single sentence or sentence pair inputs, for SuperGLUE we expand the scope to consider tasks with longer inputs. This yields a set of tasks that requires understanding individual tokens in context, complete sentences, inter-sentence relations, and entire paragraphs.

\textbf{License:} Task data must be available under licences that allow use and redistribution for research purposes.

To identify possible tasks for SuperGLUE, we disseminated a public call for task proposals to the NLP community, and received approximately 30 proposals.
We filtered these proposals according to our criteria.
Many proposals were not suitable due to licensing issues, complex formats, and insufficient headroom;
we provide examples of such tasks in Appendix~\ref{ax:excluded}. For each of the remaining tasks, we ran a BERT-based baseline and a human baseline, and filtered out tasks which were either too challenging for humans without extensive training or too easy for our machine baselines. 

\subsection{Selected Tasks}
Following this process, we arrived at eight tasks to use in SuperGLUE. See Tables \ref{tab:tasks} and \ref{tab:examples} for details and specific examples of each task.
 
\textbf{BoolQ} \citep[Boolean Questions,][]{clark2019boolq} is a QA task where each example consists of a short passage and a yes/no question about the passage. The questions are provided anonymously and unsolicited by users of the Google search engine, and afterwards paired with a paragraph from a Wikipedia article containing the answer. Following the original work, we evaluate with accuracy.

\textbf{CB} \citep[CommitmentBank,][]{demarneffe:cb} is a 
corpus of short texts in which at least one sentence contains an embedded clause. 
Each of these embedded clauses is annotated with the degree to which it appears the person who wrote the text is \textit{committed} to the truth of the clause. The resulting task framed as three-class textual entailment on examples that are drawn from the Wall Street Journal, fiction from the British National Corpus, and Switchboard.
Each example consists of a premise containing an embedded clause and the corresponding hypothesis is the extraction of that clause. 
We use a subset of the data that had inter-annotator agreement above $80\%$.
The data is imbalanced (relatively fewer \textit{neutral} examples), so we evaluate using accuracy and F1, where for multi-class F1 we compute the unweighted average of the F1 per class.

\textbf{COPA} \citep[Choice of Plausible Alternatives,][]{roemmele2011choice} is a causal reasoning task in which a system is given a premise sentence and must determine either the cause or effect of the premise from two possible choices. 
%The system must choose the alternative which has the more plausible causal relationship with the premise. 
%Examples either deal with possible \textit{causes} or %alternative possible 
%\textit{effects} of the premise sentence, accompanied by a simple question disambiguating between the two example types for the model.
All examples are handcrafted and focus on topics from blogs and a photography-related encyclopedia. Following the original work, we evaluate using accuracy.

\textbf{MultiRC} \citep[Multi-Sentence Reading Comprehension,][]{khashabi2018looking} is a QA task where each example consists of a context paragraph, a question about that paragraph, and a list of possible answers. The system must predict which answers are true and which are false. While many QA tasks exist, we use MultiRC because of a number of desirable properties: (i)~each question can have multiple possible correct answers, so each question-answer pair must be evaluated independent of other pairs, (ii)~the questions are designed such that answering each question requires drawing facts from multiple context sentences, and (iii)~the question-answer pair format more closely matches the API of other tasks in SuperGLUE than the more popular span-extractive QA format does.
The paragraphs are drawn from seven domains including news, fiction, and historical text.
The evaluation metrics are F1 over all answer-options (F1$_a$) and exact match of each question's set of answers (EM).

\textbf{ReCoRD} \citep[Reading Comprehension with Commonsense Reasoning Dataset,][]{zhang2018record} is a multiple-choice QA task.
Each example consists of a news article and a Cloze-style question about the article in which one entity is masked out. The system must predict the masked out entity from a list of possible entities in the provided passage, where the same entity may be expressed with multiple different surface forms, which are all considered correct.
Articles are from CNN and Daily Mail. We evaluate with max (over all mentions) token-level F1 and exact match (EM).

\textbf{RTE} (Recognizing Textual Entailment) datasets come from a series of annual competitions on textual entailment. %\footnote{Textual entailment is also known as natural language inference, or NLI}
RTE is included in GLUE, and we use the same data and format as GLUE: We merge data from RTE1 \citep{dagan2006pascal}, RTE2 \citep{bar2006second}, RTE3 \citep{giampiccolo2007third}, and RTE5 \citep{bentivogli2009fifth}. %\footnote{RTE4 is not publicly available, while RTE6 and RTE7 do not conform to the standard NLI task.} 
All datasets are combined and converted to two-class classification: \textit{entailment} and \textit{not\_entailment}.
Of all the GLUE tasks, RTE is among those that benefits from transfer learning the most, with performance jumping from near random-chance ($\sim$56\%) at the time of GLUE's launch to 86.3\% accuracy \citep{liu2019mt,yang2019xlnet} at the time of writing.
Given the nearly eight point gap with respect to human performance, however, the task is not yet solved by machines, and we expect the remaining gap to be difficult to close.

\textbf{WiC} \citep[Word-in-Context,][]{pilehvar2018wic} is a word sense disambiguation task cast as binary classification of sentence pairs. 
Given two text snippets and a polysemous %(sense-ambiguous) 
word that appears in both sentences, the task is to determine whether the word is used with the same sense in both sentences. 
Sentences are drawn from WordNet \citep{miller1995wordnet}, VerbNet \citep{schuler2005verbnet}, and Wiktionary.
We follow the original work and evaluate using accuracy.

\textbf{WSC} \citep[Winograd Schema Challenge,][]{levesque2012winograd} is a coreference resolution task in which examples consist of a sentence with a pronoun and a list of noun phrases from the sentence.
The system must determine the correct referrent of the pronoun from among the provided choices.
%a passage with two marked words. 
%The system must determine if the two words refer to the same thing. 
Winograd schemas are designed to require everyday knowledge and commonsense reasoning to solve.

GLUE includes a version of WSC recast as NLI, known as WNLI.
Until very recently, no substantial progress had been made on WNLI, with many submissions opting to submit majority class predictions.\footnote{WNLI is especially difficult due to an adversarial train/dev split: Premise sentences that appear in the training set often appear in the development set with a different hypothesis and a flipped label. 
If a system memorizes the training set, which was easy due to the small size of the training set, it could perform far \textit{below} chance on the development set. We remove this adversarial design in our version of WSC by ensuring that no sentences are shared between the training, validation, and test sets.}
In the past few months, several works \citep{kocijan2019surprisingly,liu2019mt} have made rapid progress via a hueristic data augmentation scheme, raising machine performance to 90.4\% accuracy.
Given estimated human performance of $\sim$96\%, there is still a gap between machine and human performance, which we expect will be relatively difficult to close.
We therefore include a version of WSC cast as binary classification, where each example consists of a sentence with a marked pronoun and noun, and the task is to determine if the pronoun refers to that noun.
The training and validation examples are drawn from the original WSC data \citep{levesque2012winograd}, as well as those distributed by the affiliated organization \textit{Commonsense Reasoning}.\footnote{\url{http://commonsensereasoning.org/disambiguation.html}} 
The test examples are derived from fiction books and have been shared with us by the authors of the original dataset.
We evaluate using accuracy.

\subsection{Scoring} As with GLUE, we seek to give a sense of aggregate system performance over all tasks by averaging scores of all tasks.
Lacking a fair criterion with which to weight the contributions of each task to the overall score, we opt for the simple approach of weighing each task equally, and for tasks with multiple metrics, first averaging those metrics to get a task score.

\subsection{Tools for Model Analysis}

\paragraph{Analyzing Linguistic and World Knowledge in Models}
GLUE includes an expert-constructed, diagnostic dataset that automatically tests models for a broad range of linguistic, commonsense, and world knowledge.
%GLUE provides an expert-constructed diagnostic dataset for automatic analysis of textual entailment output. 
Each example in this broad-coverage diagnostic is a sentence pair labeled with a three-way entailment relation (\textit{entailment}, \textit{neutral}, or \textit{contradiction})
%, matching the MultiNLI \citep{williams2018broad} label set
and tagged with labels that indicate the phenomena that characterize the relationship between the two sentences. Submissions to the GLUE leaderboard are required to include predictions from the submission's MultiNLI classifier on the diagnostic dataset, and analyses of the results were shown alongside the main leaderboard.
Since this diagnostic task has proved difficult for top models, we retain it in SuperGLUE.
However, since MultiNLI is not part of SuperGLUE, we collapse \textit{contradiction} and \textit{neutral} into a single \textit{not\_entailment} label, and request that submissions include predictions on the resulting set from the model used for the \textit{RTE} task.
We estimate human performance following the same procedure we use for the benchmark tasks (Section~\ref{ax:instruct}). %(Section~\ref{sec:human}).
We estimate an accuracy of 88\% and a Matthew's correlation coefficient (MCC, the two-class variant of the $R_3$ metric used in GLUE) of 0.77.

\paragraph{Analyzing Gender Bias in Models}
Recent work has identified the presence and amplification of many social biases in data-driven machine learning models
\citep[][ i.a.]{lu-2018-gender, zhao-2018-gender}. To promote the detection of such biases, we include Winogender \citep{rudinger-winogender} as an additional diagnostic dataset. Winogender is designed to measure gender bias in coreference resolution systems. We use the Diverse Natural Language Inference Collection \citep{poliak-2018-dnc} version that casts Winogender as a textual entailment task.%\footnote{We filter out 23 examples where the labels are ambiguous}
Each example consists of a premise sentence with a male or female pronoun and a hypothesis giving a possible antecedent of the pronoun.
Examples occur in \textit{minimal pairs}, where the only difference between an example and its pair is the gender of the pronoun in the premise.
Performance on Winogender is measured with accuracy and the \textit{gender parity score}: the percentage of minimal pairs for which the predictions are the same.
A system can trivially obtain a perfect gender parity score by guessing the same class for all examples, so a high gender parity score is meaningless unless accompanied by high accuracy.
We collect non-expert annotations to estimate human performance, and observe an accuracy of 99.7\% and a gender parity score of 0.99.

Like any diagnostic, Winogender has limitations. It offers only positive predictive value: A poor bias score is clear evidence that a model exhibits gender bias, but a good score does not mean that the model is unbiased. More specifically, in the DNC version of the task, a low gender parity score means that a model's prediction of textual entailment can be changed with a change in pronouns, all else equal. It is plausible that there are forms of bias that are relevant to target tasks of interest, but that do not surface in this setting \citep{gonen-goldberg-2019-lipstick}. 
Also, Winogender does not cover all forms of social bias, or even all forms of gender. For instance, the version of the data used here offers no coverage of gender-neutral \textit{they} or non-binary pronouns. Despite these limitations, we believe that Winogender's inclusion is worthwhile in providing a coarse sense of how social biases evolve with model performance and for keeping attention on the social ramifications of NLP models.

\section{Using SuperGLUE}

\paragraph{Software Tools}
To facilitate using SuperGLUE, we release \texttt{jiant} \citep{wang2019jiant},\footnote{\url{https://github.com/nyu-mll/jiant}} a modular software toolkit, built with PyTorch \citep{paszke2017automatic}, components from AllenNLP \citep{Gardner2017AllenNLP}, and the \texttt{transformers} package.\footnote{\url{https://github.com/huggingface/transformers}}
\texttt{jiant} implements our baselines and supports the evaluation of custom models and training methods on the benchmark tasks.
The toolkit includes support for existing popular pretrained models such as OpenAI GPT and BERT, as well as support for multistage and multitask learning of the kind seen in the strongest models on GLUE.

\paragraph{Eligibility} Any system or method that can produce predictions for the SuperGLUE tasks is eligible for submission to the leaderboard, subject to the data-use and submission frequency policies stated immediately below. There are no restrictions on the type of methods that may be used, and there is no requirement that any form of parameter sharing or shared initialization be used across the tasks in the benchmark. 
To limit overfitting to the private test data, users are limited to a maximum of two submissions per day and six submissions per month.

\paragraph{Data} 
Data for the tasks are available for download through the SuperGLUE site and through a download script included with the software toolkit. 
Each task comes with a standardized training set, development set, and \textit{unlabeled} test set.
Submitted systems may use any public or private data when developing their systems, with a few exceptions: Systems may only use the SuperGLUE-distributed versions of the task datasets, as these use different train/validation/test splits from other public versions in some cases. Systems also may not use the unlabeled test data for the tasks in system development in any way, may not use the structured source data that was used to collect the WiC labels (sense-annotated example sentences from WordNet, VerbNet, and Wiktionary) in any way, and may not build systems that share information across separate \textit{test} examples in any way.

%We do not endorse the use of the benchmark data for \textit{non-research} applications, due to concerns about socially relevant biases (such as ethnicity--occupation associations) that may be undesirable or legally problematic in deployed systems. Because these biases are evident in texts from a wide variety of sources and collection methods \citep[e.g., ][]{rudinger2017social}, and because none of our task datasets directly mitigate them, one can reasonably presume that our training sets teach models these biases to some extent and that our evaluation sets similarly \textit{reward} models that learn these biases. 

%\paragraph{The Leaderboard} To compete on the benchmark, authors must submit a zip file containing predictions from their system to the website to be scored by an auto-grader. By default, all submissions are private. %; submissions which are marked public will be placed on appropriate position on the leaderboard based on their score. 
%To submit a system to the public leaderboard, one must score it and fill out a short additional form supplying either a short description or a link to a paper. Anonymous submissions are allowed, but will only be posted only when they are accompanied by an (anonymized) full paper. Users are limited to a maximum of two submissions per day and six submissions per month.

To ensure reasonable credit assignment, because we build very directly on prior work, we ask the authors of submitted systems to directly name and cite the specific datasets that they use, \textit{including the benchmark datasets}. We will enforce this as a requirement for papers to be listed on the leaderboard.

\section{Experiments}

\subsection{Baselines}

%\paragraph{Model}
\paragraph{BERT}

\begin{table*}[t]
\footnotesize

\caption{Baseline performance on the SuperGLUE test sets and diagnostics. 
For CB we report accuracy and macro-average F1. For MultiRC we report F1 on all answer-options and exact match of each question's set of correct answers. AX$_b$ is the broad-coverage diagnostic task, scored using Matthews' correlation (MCC). AX$_g$ is the Winogender diagnostic, scored using accuracy and the gender parity score (GPS). All values are scaled by 100. The \textit{Avg} column is the overall benchmark score on non-AX$_*$ tasks.
The bolded numbers reflect the best machine performance on task.
*MultiRC has multiple test sets released on a staggered schedule, 
%*MultiRC uses a staggered test set release, 
and these results evaluate on an installation of the test set that is a subset of ours.
%We show results on the standardized development sets in Table \todo{N} in the appendix.} %The $\Delta$ row reports the difference between the human performance baseline and BERT-Large.
}

\centering %\small 
\fontsize{8.4}{10.1}\selectfont
\setlength{\tabcolsep}{0.3em}

%\begin{tabular}{lcc@{/}ccc@{/}cccc}
\begin{tabular}{lccc@{/}ccc@{/}cc@{/}cccccc@{/}c}

\toprule

% && \multicolumn{4}{c}{\textbf{Word Level}} & \multicolumn{3}{c}{\textbf{Sentence Level}} & \textbf{Paragraph Level} \\

\textbf{Model} & \multicolumn{1}{c}{\textbf{Avg}} & \multicolumn{1}{c}{\textbf{BoolQ}} & \multicolumn{2}{c}{\textbf{CB}} &  \multicolumn{1}{c}{\textbf{COPA}} & \multicolumn{2}{c}{\textbf{MultiRC}} & \multicolumn{2}{c}{\textbf{ReCoRD}} & \multicolumn{1}{c}{\textbf{RTE}} & \multicolumn{1}{c}{\textbf{WiC}} & \multicolumn{1}{c}{\textbf{WSC}} & 
\multicolumn{1}{c}{\textbf{AX$_b$}} &
\multicolumn{2}{c}{\textbf{AX$_g$}} \\
%\midrule
\textbf{Metrics} & \multicolumn{1}{c}{} & \multicolumn{1}{c}{\textbf{Acc.}} & \multicolumn{2}{c}{\textbf{F1/Acc.}} &  \multicolumn{1}{c}{\textbf{Acc.}} & \multicolumn{2}{c}{\textbf{F1$_a$/EM}} & \multicolumn{2}{c}{\textbf{F1/EM}} & \multicolumn{1}{c}{\textbf{Acc.}} & \multicolumn{1}{c}{\textbf{Acc.}} & \multicolumn{1}{c}{\textbf{Acc.}} &
\multicolumn{1}{c}{\textbf{MCC}} & 
\multicolumn{1}{c}{\textbf{GPS}} & 
\multicolumn{1}{c}{\textbf{Acc.}} \\
\midrule
% GAP outisde best: 72.1 & 98.0

% v1.0
%Most Frequent Class & 47.1 & 62.3 & 48.4 & 21.7 & 50 & 61.1 & 0.3 & 33.4 & 32.5 & 50.4 & 50.0 & 65.1 & 0.0 & & 100.0 \\
%CBOW & 48.6 & & 69.2 & 47.6 & 52.2 & 38.8 & 0.5 & 23.0 & 22.3 & 50.4 & 50.0 & 61.0 & & & \\
%BERT & 68.0 & & 84.4 & 80.6 & 69.8 & 66.2 & 22.2 & 67.6 & 66.9 & 73.2 & \textbf{70.4} & \textbf{65.1} & & & \\
%BERT++ & \textbf{70.5} & & \textbf{88.4} & \textbf{82.7} & 77.4 & 66.2 & 22.2 & 67.7 & 66.9 & 77.6 & \textbf{70.4} & \textbf{65.1} & & & \\

Most Frequent & 47.1 & 62.3 & 21.7 & 48.4 & 50.0 & 61.1 & 0.3 & 33.4 & 32.5 & 50.3 & 50.0 & 65.1 & 0.0 & 100.0 & 50.0 \\
CBoW & 44.3 & 62.1 & 49.0  & 71.2 &  51.6 & 0.0 & 0.4 & 14.0 & 13.6 & 49.7 & 53.0 & 65.1 & -0.4 & 100.0 & 50.0 \\
BERT & 69.0 & 77.4 & 75.7 & 83.6 & 70.6 & 70.0 & 24.0 & 72.0 & 71.3 & 71.6 & \textbf{69.5} & \textbf{64.3} & 23.0 & 97.8 & 51.7 \\
BERT++ & \textbf{71.5} & 79.0 & \textbf{84.7} & \textbf{90.4} & 73.8 & 70.0 & 24.1 & 72.0 & 71.3 & 79.0 & \textbf{69.5} & \textbf{64.3} & 38.0 & 99.4 & 51.4 \\

% sources:
% CB: our own (w/ MNLI intermediate training)
% CoPA: GPT
% GAP:
% MultiRC: https://arxiv.org/pdf/1902.08852.pdf (70.4 F1a / 24.5 EM) but evaluates on a subset of our test set
% RTE: MT-DNN (GLUE leaderboard)
% WiC: our own (WiC CodaLab page)
% WSC: our own (since this is evaluating each prompt-answer pair?)
Outside Best & - & \textbf{80.4} & - &  - & \textbf{84.4} & \textbf{70.4}* & \textbf{24.5}* & \textbf{74.8} & \textbf{73.0} & \textbf{82.7} & - & - & - & - & - \\

\midrule
% most frequent baseline bias number = 0.9638; human = 0.992647
Human (est.) & 89.8 & 89.0 & 95.8 & 98.9 & 100.0 & 81.8* & 51.9* & 91.7 & 91.3 & 93.6 & 80.0 & 100.0 & 77.0 & 99.3 & 99.7 \\
% avg = 92.3 w/ F1s metric for MultiRC

\bottomrule
\end{tabular}

\label{tab:benchmark}
\end{table*}

Our main baselines are built around BERT, variants of which are among the most successful approach on GLUE at the time of writing. Specifically, we use the \texttt{bert-large-cased} variant.
Following the practice recommended in \citet{devlin2018bert}, for each task, we use the simplest possible architecture on top of BERT. We fine-tune a copy of the pretrained BERT model separately for each task, and leave the development of multi-task learning models to future work. For training, we use the procedure specified in \citet{devlin2018bert}:
We use Adam \citep{kingma2014adam} with an initial learning rate of $10^{-5}$ and fine-tune for a maximum of 10 epochs.

For classification tasks with sentence-pair inputs (BoolQ, CB, RTE, WiC), we concatenate the sentences with a \textsc{[sep]} token, feed the fused input to BERT, and use a logistic regression classifier that sees the representation corresponding to \textsc{[cls]}.
For WiC, we also concatenate the representation of the marked word. % to the \textsc{[cls]} representation.
For COPA, MultiRC, and ReCoRD, for each answer choice, we similarly concatenate the context with that answer choice and feed the resulting sequence into BERT to produce an answer representation.
For COPA, we project these representations into a scalar, and take as the answer the choice with the highest associated scalar.
For MultiRC, because each question can have more than one correct answer, we feed each answer representation into a logistic regression classifier.
For ReCoRD, we also evaluate the probability of each candidate independent of other candidates, and take the most likely candidate as the model's prediction.
For WSC, which is a span-based task, we use a model inspired by \citet{tenney2018you}.
Given the BERT representation for each word in the original sentence, we get span representations of the pronoun and noun phrase via a self-attention span-pooling operator \citep{lee2017end}, before feeding it into a logistic regression classifier. 
%See Appendix \ref{ax:baselines} for training details.

\paragraph{BERT++} We also report results using BERT with additional training on related datasets before fine-tuning on the benchmark tasks, following the STILTs style of transfer learning \citep{phang2018sentence}. Given the productive use of MultiNLI in pretraining and intermediate fine-tuning of pretrained language models \citep[][i.a.]{conneau2017supervised,phang2018sentence}, for CB, RTE, and BoolQ, we use MultiNLI as a transfer task by first using the above procedure on MultiNLI. Similarly, given the similarity of COPA to SWAG \citep{zellers2018swag}, we first fine-tune BERT on SWAG. These results are reported as BERT++. For all other tasks, we reuse the results of BERT fine-tuned on just that task.

\paragraph{Other Baselines} We include a baseline where for each task we simply predict the majority class,\footnote{For ReCoRD, we predict the entity that has the highest F1 with the other entity options.} as well as a bag-of-words baseline where each input is represented as an average of its tokens' GloVe word vectors \citep[the 300D/840B release from][]{pennington2014glove}. 
%\paragraph{Outside Best} 
Finally, we list the best known result on each task as of May 2019, except on tasks which we recast (WSC), resplit (CB), or achieve the best known result (WiC). 
The outside results for COPA, MultiRC, and RTE are from  \citet{sap2019socialiqa}, \citet{trivedi2019repurposing}, and \citet{liu2019mt} respectively.

%\subsection{Human Performance}\label{sec:human}
%Several datasets have non-expert human performance baselines already available. 
\paragraph{Human Performance}
\citet{pilehvar2018wic}, \citet{khashabi2018looking}, \citet{nangia2019human}, and \citet{zhang2018record} respectively provide estimates for human performance on WiC, MultiRC, RTE, and ReCoRD. 
For the remaining tasks, including the diagnostic set, we estimate human performance by hiring crowdworker annotators through Amazon's Mechanical Turk platform to reannotate a sample of each test set.
%\paragraph{Training Phase} 
We follow a two step procedure where a crowd worker completes a short training phase before proceeding to the annotation phase, modeled after the method used by \citet{nangia2019human} for GLUE. 
%For both phases and all tasks, the average pay rate is \$23.75/hr. \footnote{This estimate is taken from \url{https://turkerview.com}.}
%, where crowd workers self-report their hourly income on tasks.}
See Appendix~\ref{ax:instruct} for details.

%The training phase uses 30 examples taken from the development set of the task. 
%In the training phase, workers are provided with instructions on the task, linked to an FAQ page, and are asked to annotate up to 30 examples from the development set. After answering each example, workers are also asked to check their work against the provided ground truth label.
%by clicking on a ``Check Work" button which reveals the ground truth label. 
%\paragraph{Annotation Phase} 
%After the training phase is complete, we provide the qualification to work on the annotation phase to all workers who annotated a minimum of five examples, i.e. completed five HITs during training and achieved performance at, or above the median performance across all workers during training. 

%In the annotation phase, workers are provided with the same instructions as the training phase, and are linked to the same FAQ page. The instructions for all tasks are provided in Appendix~\ref{ax:instruct}.
%For the annotation phase we randomly sample 100 examples from the task's test set, with the exception of WSC where we annotate the full test set. For each  example, we collect annotations from five workers and take a majority vote to estimate human performance.
%For additional details, see Appendix~\ref{ax:voting}.
%For task-specific details on how we present the tasks to annotators and calculate human performance numbers, refer to Appendix~\ref{ax:voting}.

\subsection{Results}

Table~\ref{tab:benchmark} shows results for all baselines. 
The most frequent class and CBOW baselines do not perform well overall, achieving near chance performance for several of the tasks.
Using BERT increases the average SuperGLUE score by 25 points,
attaining significant gains on all of the benchmark tasks, particularly MultiRC, ReCoRD, and RTE. On WSC, BERT actually performs worse than the simple baselines, likely due to the small size of the dataset and the lack of data augmentation.
%On CB, we achieve strong accuracy and F1 scores of 84.4 and 80.6 respectively. 
Using MultiNLI as an additional source of supervision for BoolQ, CB, and RTE leads to a 2-5 point improvement on all tasks.
Using SWAG as a transfer task for COPA sees an 8 point improvement.

Our best baselines still lag substantially behind human performance.
On average, there is a nearly 20 point gap between \textsc{BERT++} and human performance. The largest gap is on WSC, with a 35 point difference between the best model and human performance. 
The smallest margins are on BoolQ, CB, RTE, and WiC, with gaps of around 10 points on each of these.
We believe these gaps will be challenging to close: On WSC and COPA, human performance is perfect. On three other tasks, it is in the mid-to-high 90s. 
%We believe this a reflection of the fact that current state-of-the-art models, like BERT, are genuinely fairly effective at sentence understanding in non-adversarial settings.
On the diagnostics, all models continue to lag significantly behind humans.
Though all models obtain near perfect gender parity scores on Winogender, this is due to the fact that they are obtaining accuracy near that of random guessing.

\section{Conclusion}

We present SuperGLUE, a new benchmark for evaluating general-purpose language understanding systems.
SuperGLUE updates the GLUE benchmark by identifying a new set of challenging NLU tasks, as measured by the difference between human and machine baselines.
The set of eight tasks in our benchmark emphasizes diverse task formats and low-data training data tasks, with nearly half the tasks having fewer than 1k examples and all but one of the tasks having fewer than 10k examples.

We evaluate BERT-based baselines and find that they still lag behind humans by nearly 20 points.
Given the difficulty of SuperGLUE for BERT, we expect that further progress in multi-task, transfer, and unsupervised/self-supervised learning techniques will be necessary to approach human-level performance on the benchmark.
Overall, we argue that SuperGLUE offers a rich and challenging testbed for work developing new general-purpose machine learning methods for language understanding.

\section{Acknowledgments}

We thank the original authors of the included datasets in SuperGLUE for their cooperation in the creation of the benchmark, as well as those who proposed tasks and datasets that we ultimately could not include.
This work was made possible in part by a donation to NYU from Eric and Wendy Schmidt made by recommendation of the Schmidt Futures program. We gratefully acknowledge the support of the NVIDIA Corporation with the donation of a Titan V GPU used at NYU for this research, and funding from DeepMind for the hosting of the benchmark platform.
AW is supported by the National Science Foundation Graduate Research Fellowship Program under Grant No. DGE 1342536. Any opinions, findings, and conclusions or recommendations expressed in this material are those of the author(s) and do not necessarily reflect the views of the National Science Foundation.
This project is partly supported by Samsung Advanced Institute of Technology (Next Generation Deep Learning: from Pattern Recognition to AI) and Samsung Electronics (Improving Deep Learning using Latent Structure).

\bibliography{neurips_2019}
\bibliographystyle{plainnat}

\clearpage

% \section*{\Large{Supplementary Material for A Stickier Benchmark for General-Purpose Language Understanding Systems}}
\appendix

\section{Development Set Results}\label{ax:dev}

In Table~\ref{tab:benchmark-dev}, we present results of the baselines on the SuperGLUE tasks development sets.

\begin{table*}[t]

\caption{Baseline performance on the SuperGLUE development.}

\centering \fontsize{8.4}{10.1}\selectfont \setlength{\tabcolsep}{0.5em}

%\begin{tabular}{lcr@{/}lcr@{/}lr@{/}lccc}
\begin{tabular}{lccc@{/}ccc@{/}cc@{/}cccc}

\toprule

\textbf{Model} & \multicolumn{1}{c}{\textbf{Avg}} & \multicolumn{1}{c}{\textbf{BoolQ}} & \multicolumn{2}{c}{\textbf{CB}} &  \multicolumn{1}{c}{\textbf{COPA}} & \multicolumn{2}{c}{\textbf{MultiRC}} & \multicolumn{2}{c}{\textbf{ReCoRD}} & \multicolumn{1}{c}{\textbf{RTE}} & \multicolumn{1}{c}{\textbf{WiC}} & \multicolumn{1}{c}{\textbf{WSC}} \\ %\multicolumn{1}{c}{\textbf{AX}} \\
%\midrule
\textbf{Metrics} & \multicolumn{1}{c}{} & \multicolumn{1}{c}{\textbf{Acc.}} & \multicolumn{2}{c}{\textbf{Acc./F1}} &  \multicolumn{1}{c}{\textbf{Acc.}} & \multicolumn{2}{c}{\textbf{F1$_a$/EM}} & \multicolumn{2}{c}{\textbf{F1/EM}} & \multicolumn{1}{c}{\textbf{Acc.}} & \multicolumn{1}{c}{\textbf{Acc.}} & \multicolumn{1}{c}{\textbf{Acc.}} \\
\midrule
% most frequent baseline bias number = 0.9638; human = 0.992647

Most Frequent Class & 47.7 & 62.2 & 50.0 & 22.2 & 55.0 & 59.9 & 0.8 & 32.4 & 31.5 & 52.7 & 50.0 & 63.5 \\

CBOW & 47.7 & 62.4 & 71.4 & 49.6 & 63.0 & 20.3 & 0.3 & 14.4 & 13.8 & 54.2 & 55.3 & 61.5 \\
%CBOW & & 54.4 & 58.2 & 100.4 & 62.2 & 49.6 & 54.1 & 69.2 & 47.6 & & 41.2 \\

BERT & 72.2 & 77.7 & 94.6 & 93.7 & 69.0 & 70.5 & 24.7 & 70.6 & 69.8 & 75.8 & 74.9 & 68.3 \\
%BERT & 64.9 & & 87.5 & 88.4 & 72.0 & 69.2 & 23.8 & & & 75.2 & 74.1 &  61.5  \\

BERT++ & 74.6 & 80.1 & 96.4 & 95.0 & 78.0 & 70.5 & 24.7 & 70.6 & 69.8 & 82.3 & 74.9 & 68.3 \\
%BERT++ & 66.1 & & 89.3 & 81.5 & 74.0 & 69.2 & 23.8 & & & 81.9 & 75.2 &  63.5 \\

\bottomrule
\end{tabular}

\label{tab:benchmark-dev}
\end{table*}

\section{Performance on GLUE Diagnostics}\label{ax:diagnostics}
\begin{figure}[!h]
    \centering
    \includegraphics[width=0.8\textwidth]{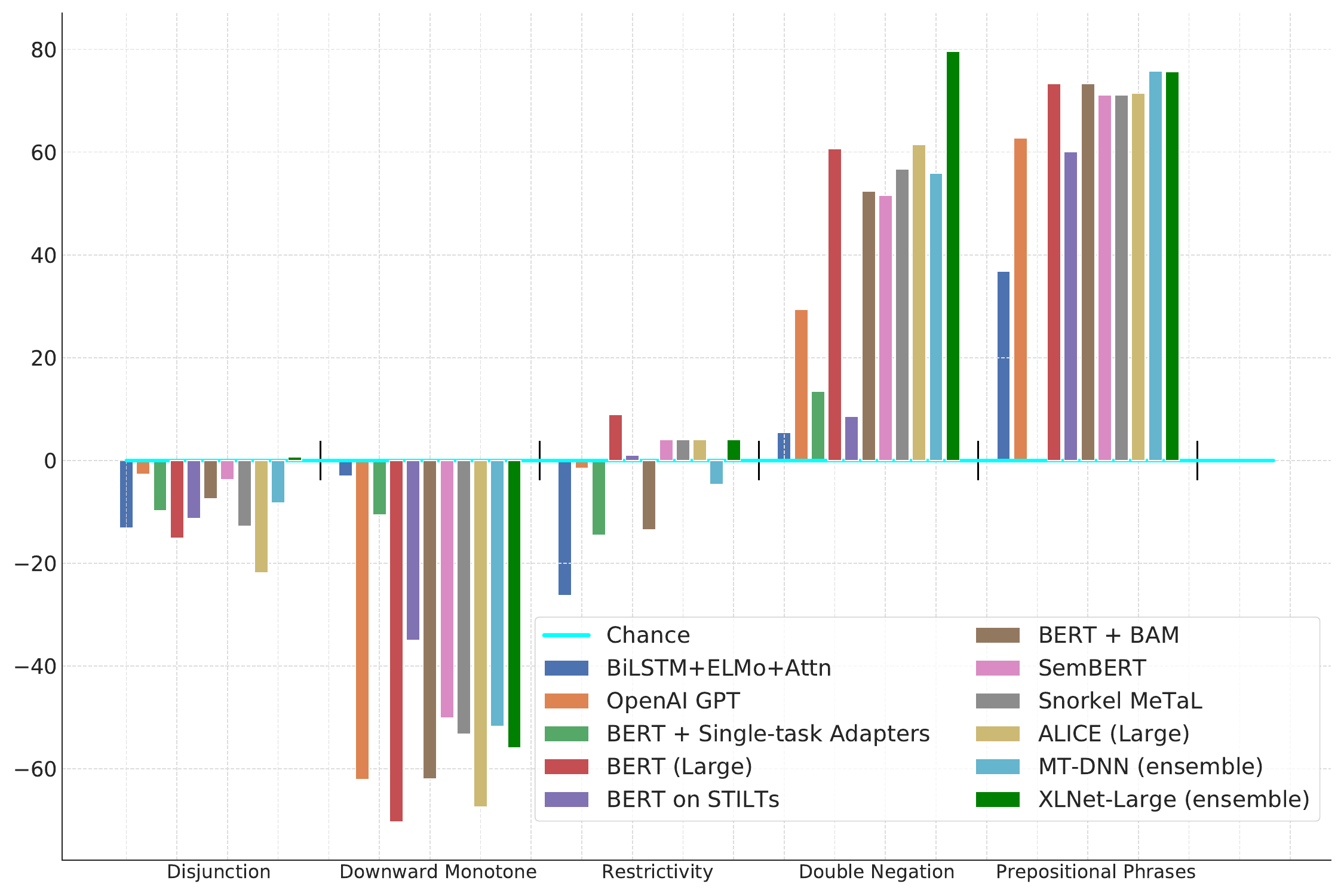}
    % diagnostic trends plot
    \caption{Performance of GLUE submissions on selected diagnostic categories, reported using the $R_3$ metric scaled up by 100, as in \citet[see paper for a description of the categories]{wang2018glue}.
    Some initially difficult categories, like double negation, saw gains from advances on GLUE, but others remain hard (restrictivity) or even adversarial (disjunction, downward monotone).}
    
    \label{fig:diagnostic-trends}
\end{figure}

Figure~\ref{fig:diagnostic-trends} shows the performance on the GLUE diagnostics dataset for systems submitted to the public leaderboard.

\section{Human Performance Estimation}\label{ax:instruct}

For collecting data to establish human performance on the SuperGLUE tasks, we follow a two step procedure where we first provide some training to the crowd workers before they proceed to annotation. 
%The training phase uses 30 examples taken from the development set of the task. 
For both steps and all tasks, the average pay rate is \$23.75/hr.\footnote{This estimate is taken from \url{https://turkerview.com}.}

In the training phase, workers are provided with instructions on the task, linked to an FAQ page, and are asked to annotate up to 30 examples from the development set. After answering each example, workers are also asked to check their work against the provided ground truth label.
%by clicking on a ``Check Work" button which reveals the ground truth label. 
%\paragraph{Annotation Phase} 
After the training phase is complete, we provide the qualification to work on the annotation phase to all workers who annotated a minimum of five examples, i.e. completed five HITs during training and achieved performance at, or above the median performance across all workers during training. 

In the annotation phase, workers are provided with the same instructions as the training phase, and are linked to the same FAQ page. The instructions for all tasks are provided in Appendix~\ref{ax:instruct}.
For the annotation phase we randomly sample 100 examples from the task's test set, with the exception of WSC where we annotate the full test set. For each example, we collect annotations from five workers and take a majority vote to estimate human performance.
For additional details, see Appendix~\ref{ax:voting}.

\subsection{Training Phase Instructions}\label{ax:traininstruct}
In the training step, we provide workers with brief instructions about the training phase. An example of these instructions is given Table~\ref{tab:training}. These training instructions are the same across tasks, only the task name in the instructions is changed.

\subsection{Task Instructions}\label{ax:testinstruct}
During training and annotation for each task, we provide workers with brief instructions tailored to the task. We also link workers to an FAQ page for the task. Tables~\ref{tab:copa}, \ref{tab:cb}, \ref{tab:wsc}, and \ref{tab:boolq0}, show the instructions we used for all four tasks: COPA, CommitmentBank, WSC, and BoolQ respectively. The instructions given to crowd workers for annotations on the diagnostic and bias diagnostic datasets are shown in Table~\ref{tab:ax}.

We collected data to produce conservative estimates for human performance on several tasks that we did not ultimately include in our benchmark, including GAP \citep{webster2018mind}, PAWS \citep{zhang2019paws}, Quora Insincere Questions,\footnote{\url{https://www.kaggle.com/c/quora-insincere-questions-classification/data}} Ultrafine Entity Typing \citep{choi2018ultra}, and Empathetic Reactions datasets \citep{Buechel18emnlp}. The instructions we used for these tasks are shown in Tables~\ref{tab:gap}, \ref{tab:paws},
\ref{tab:quora}, \ref{tab:ultra}, and \ref{tab:empathy}. 

% \paragraph{Ultrafine Entity Typing}\label{ax:ultrafine} We cast the task into a binary classification problem to make it an easier task for non-expert crowd workers. We work in cooperation with the authors of the dataset \citep{choi2018ultra} to do this reformulation: We give workers one possible tag for a word or phrase and asked them to classify the tag as being applicable or not. 

% The authors used WordNet \citep{miller1995wordnet} to expand the set of labels to include synonyms and hypernyms from WordNet. They then asked five annotators to validate these tags. The tags from this validation had high agreement, and were included in the publicly available Ultrafine Entity Typing dataset,\footnote{\url{https://homes.cs.washington.edu/~eunsol/open_entity.html}} This constitutes our set of positive examples. The rest of the tags from the validation procedure that are not in the public dataset constitute our negative examples.

% \paragraph{GAP} For the Gendered Ambiguous Pronoun Coreference task \citep[GAP,][]{webster2018mind}, we simplified the task by providing noun phrase spans as part of the input, thus reducing the original structure prediction task to a classification task. This task was presented to crowd workers as a three way classification problem: Choose span A, B, or neither. 

\subsection{Task Specific Details}\label{ax:voting}
For WSC and COPA we provide annotators with a two way classification problem. We then use majority vote across annotations to calculate human performance.

\paragraph{CommitmentBank} We follow the authors in providing annotators with a 7-way classification problem. We then collapse the annotations into 3 classes by using the same ranges for bucketing used by \cite{demarneffe:cb}. We then use majority vote to get human performance numbers on the task. 

Furthermore, for training on CommitmentBank we randomly sample examples from the low inter-annotator agreement portion of the CommitmentBank data that is not included in the benchmark version of the task. These low agreement examples are generally harder to classify since they are more ambiguous.

\paragraph{Diagnostic Dataset} Since the diagnostic dataset does not come with accompanying training data, we train our workers on examples from RTE's development set. RTE is also a textual entailment task and is the most closely related task in the main benchmark. Providing the crowd workers with training on RTE enables them to learn label definitions which should generalize to the diagnostic dataset.

\paragraph{Ultrafine Entity Typing}\label{ax:ultrafine} We cast the task into a binary classification problem to make it an easier task for non-expert crowd workers. We work in cooperation with the authors of the dataset \citep{choi2018ultra} to do this reformulation: We give workers one possible tag for a word or phrase and asked them to classify the tag as being applicable or not. 

The authors used WordNet \citep{miller1995wordnet} to expand the set of labels to include synonyms and hypernyms from WordNet. They then asked five annotators to validate these tags. The tags from this validation had high agreement, and were included in the publicly available Ultrafine Entity Typing dataset,\footnote{\url{https://homes.cs.washington.edu/~eunsol/open_entity.html}} This constitutes our set of positive examples. The rest of the tags from the validation procedure that are not in the public dataset constitute our negative examples.

\paragraph{GAP} For the Gendered Ambiguous Pronoun Coreference task \citep[GAP,][]{webster2018mind}, we simplified the task by providing noun phrase spans as part of the input, thus reducing the original structure prediction task to a classification task. This task was presented to crowd workers as a three way classification problem: Choose span A, B, or neither.

\section{Excluded Tasks}\label{ax:excluded}

In this section we provide some examples of tasks that we evaluated for inclusion but ultimately could not include. We report on these excluded tasks only with the permission of their authors. 
We turned down many medical text datasets because they are usually only accessible with explicit permission and credentials from the data owners. 

Tasks like QuAC \citep{choi2018quac} and STREUSLE \citep{schneider2015corpus} differed substantially from the format of other tasks in our benchmark, which we worried would incentivize users to spend significant effort on task-specific model designs, rather than focusing on general-purpose techniques.
It was challenging to train annotators to do well on Quora Insincere Questions \footnote{\url{https://www.kaggle.com/c/quora-insincere-questions-classification/data}}, 
Empathetic Reactions \citep{Buechel18emnlp}, and a recast version of Ultra-Fine Entity Typing \citep[][see Appendix \ref{ax:ultrafine} for details]{choi2018ultra}, leading to low human performance.
BERT achieved very high or superhuman performance on Query Well-Formedness \citep{faruqui2018identifying}, PAWS \citep{zhang2019paws}, Discovering Ongoing Conversations \citep{zanzotto2017have}, and GAP \citep{webster2018mind}.

During the process of selecting tasks for our benchmark, we collected human performance baselines and run BERT-based machine baselines for some tasks that we ultimately excluded from our task list. We chose to exclude these tasks because our BERT baseline performs better than our human performance baseline or if the gap between human and machine performance is small. 

On Quora Insincere Questions our BERT baseline outperforms our human baseline by a small margin: an F1 score of 67.2 versus 66.7 for BERT and human baselines respectively. Similarly, on the Empathetic Reactions dataset, BERT outperforms our human baseline, where BERT's predictions have a Pearson correlation of 0.45 on empathy and 0.55 on distress, compared to 0.45 and 0.35 for our human baseline. For PAWS-Wiki, we report that BERT achieves an accuracy of 91.9\%, while our human baseline achieved 84\% accuracy.
These three tasks are excluded from the benchmark since our, admittedly conservative, human baselines are worse than machine performance. Our human performance baselines are subject to the clarity of our instructions (all instructions can be found in Appendix~\ref{ax:instruct}), and crowd workers engagement and ability.

For the Query Well-Formedness task, the authors set an estimate human performance at 88.4\% accuracy. Our BERT baseline model reaches an accuracy of 82.3\%. While there is a positive gap on this task, the gap was smaller than we were were willing to tolerate.
Similarly, on our recast version of the Ultrafine Entity Typing, we observe too small a gap between human (60.2 F1) and machine performance (55.0 F1). Our recasting for this task is described in Appendix~\ref{ax:testinstruct}.
On GAP, when taken as a classification problem without the related task of span selection (details in \ref{ax:testinstruct}), BERT performs (91.0 F1) comparably to our human baseline (94.9 F1). Given this small margin, we also exclude GAP.

On Discovering Ongoing Conversations, our BERT baseline achieves an F1 of 51.9 on a version of the task cast as sentence pair classification (given two snippets of texts from plays, determine if the second snippet is a continuation of the first). This dataset is very class imbalanced (90\% negative), so we also experimented with a class-balanced version on which our BERT baselines achieves 88.4 F1. 
Qualitatively, we also found the task challenging for humans as there was little context for the text snippets and the examples were drawn from plays using early English.
Given this fairly high machine performance and challenging nature for humans,  we exclude this task from our benchmark.\\~\\~\\

\noindent
\textit{Instructions tables begin on the following page.}

\clearpage

\begin{table}[!t]
\caption{The instructions given to crowd-sourced worker describing the training phase for the Choice of Plausible Answers (COPA) task.\vspace{0.2cm}}
\begin{tabular}{p{0.95\textwidth}}
\toprule
\vspace{0.1cm}
The New York University Center for Data Science is collecting your answers for use in research on computer understanding of English. Thank you for your help!\\\\

This project is a \textbf{training task} that needs to be completed before working on the main project on AMT named Human Performance: Plausible Answer. Once you are done with the training, please proceed to the main task! The qualification approval is not immediate but we will add you to our qualified workers list within a day.\\\\

In this training, you must answer the question on the page and then, to see how you did, click the \textbf{Check Work} button at the bottom of the page before hitting Submit. The Check Work button will reveal the true label. Please use this training and the provided answers to build an understanding of what the answers to these questions look like (the main project, Human Performance: Plausible Answer, does not have the answers on the page).\bigskip\\
\bottomrule
\end{tabular}
\label{tab:training}
\end{table}

%%%%%%%%%%%%%%%%%%%%%%%%%%%%%%%%%%%%%%%%%%%%%
%%%%%%%%%%%%%%% CoPA %%%%%%%%%%%%%%%
%%%%%%%%%%%%%%%%%%%%%%%%%%%%%%%%%%%%%%%%%%%%%
\begin{table}[!t]
\caption{Task-specific instructions for Choice of Plausible Alternatives (COPA). These instructions were provided during both training and  annotation phases.\vspace{0.2cm}}
\begin{tabular}{p{0.95\textwidth}}
\toprule
\vspace{0.1cm}
\textbf{Plausible Answer Instructions}\\\\
The New York University Center for Data Science is collecting your answers for use in research on computer understanding of English. Thank you for your help!\\\\

We will present you with a prompt sentence and a question. The question will either be about what caused the situation described in the prompt, or what a possible effect of that situation is. We will also give you two possible answers to this question. \textcolor{red}{Your job is to decide, given the situation described in the prompt, which of the two options is a more plausible answer to the question:}\\\\

In the following example, option \underline{1.} is a more plausible answer to the question about what caused the situation described in the prompt,\\

\begin{quote}
\textit{\textcolor{mygreen}{The girl received a trophy.}}

\textit{What's the CAUSE for this?}
\textcolor{mypurple}{
\begin{enumerate}
    \item \textit{She won a spelling bee.}
    \item \textit{She made a new friend.}
\end{enumerate}}
\end{quote}

\medskip
In the following example, option \underline{2.} is a more plausible answer the question about what happened because of the situation described in the prompt,
\begin{quote}
\textit{\textcolor{mygreen}{The police aimed their weapons at the fugitive.}}

\textit{What happened as a RESULT?}
\textcolor{mypurple}{
\begin{enumerate}
    \item \textit{The fugitive fell to the ground.}
    \item \textit{The fugitive dropped his gun.}
\end{enumerate}}
\end{quote}

\medskip
If you have any more questions, please refer to our \underline{\textcolor{blue}{\href{https://nyu-mll.github.io/SuperGLUE-human/copa-faq}{FAQ}}} page.\medskip\\
\bottomrule
\end{tabular}
\label{tab:copa}
\end{table}

%%%%%%%%%%%%%%%%%%%%%%%%%%%%%%%%%%%%%%%%%%%%%
%%%%%%%%%%%%%%% Commitment Bank %%%%%%%%%%%%%%%
%%%%%%%%%%%%%%%%%%%%%%%%%%%%%%%%%%%%%%%%%%%%%

\begin{table}[!t]
\caption{Task-specific instructions for Commitment Bank. These instructions were provided during both training and  annotation phases.\vspace{0.2cm}}
\begin{tabular}{p{0.95\textwidth}}
\toprule
\vspace{0.1cm}
\textbf{Speaker Commitment Instructions}\\\\
The New York University Center for Data Science is collecting your answers for use in research on computer understanding of English. Thank you for your help!\\\\

We will present you with a prompt taken from a piece of dialogue, this could be a single sentence, a few sentences, or a short exchange between people. \textcolor{red}{Your job is to figure out, based on this first prompt (on top), how certain the speaker is about the truthfulness of the second prompt (on the bottom).} You can choose from a 7 point scale ranging from (1) completely certain that the second prompt is true to (7) completely certain that the second prompt is false. Here are examples for a few of the labels:\\\\

Choose \underline{1 (certain that it is true)} if the speaker from the first prompt definitely believes or knows that the second prompt is true. For example,

\begin{quote}
\textit{\textcolor{mygreen}{"What fun to hear Artemis laugh. She’s such a serious child. I didn’t know she had a sense of humor."}}

\textit{\textcolor{mypurple}{"Artemis had a sense of humor"}}
\end{quote}

\medskip

Choose \underline{4 (not certain if it is true or false)} if the speaker from the first prompt is uncertain if the second prompt is true or false. For example,
\begin{quote}
\textit{\textcolor{mygreen}{"Tess is committed to track. She’s always trained with all her heart and soul. One can only hope that she has recovered from the flu and will cross the finish line."}}

\textit{\textcolor{mypurple}{"Tess crossed the finish line."}}
\end{quote}

\medskip

Choose \underline{7 (certain that it is false)} if the speaker from the first prompt definitely believes or knows that the second prompt is false. For example,
\begin{quote}
\textit{\textcolor{mygreen}{"Did you hear about Olivia’s chemistry test? She studied really hard. But even after putting in all that time and energy, she didn’t manage to pass the test". }}

\textit{\textcolor{mypurple}{"Olivia passed the test."}}
\end{quote}

\medskip
If you have any more questions, please refer to our \underline{\textcolor{blue}{\href{https://nyu-mll.github.io/SuperGLUE-human/commit-faq}{FAQ}}} page.\medskip\\
\bottomrule
\end{tabular}
\label{tab:cb}
\end{table}

%%%%%%%%%%%%%%%%%%%%%%%%%%%%%%%%%%%%%%%%%%%%%
%%%%%%%%%%%%%%% Winograd Schema Challenge %%%%%%%%%%%%%%%
%%%%%%%%%%%%%%%%%%%%%%%%%%%%%%%%%%%%%%%%%%%%%

\begin{table}[!t]
\caption{Task-specific instructions for Winograd Schema Challenge (WSC). These instructions were provided during both training and  annotation phases.\vspace{0.2cm}}
\begin{tabular}{p{0.95\textwidth}}
\toprule
\vspace{0.1cm}
\textbf{Winograd Schema Instructions}\\\\
The New York University Center for Data Science is collecting your answers for use in research on computer understanding of English. Thank you for your help!\\\\

We will present you with a sentence that someone wrote, with one bolded pronoun. We will then ask if you if the pronoun refers to a specific word or phrase in the sentence. \textcolor{red}{Your job is to figure out, based on the sentence, if the bolded pronoun refers to this selected word or phrase:}\\ \\

Choose \underline{Yes} if the pronoun refers to the selected word or phrase. For example,

\begin{quote}
\textit{\textcolor{mygreen}{"I put the cake away in the refrigerator. It has a lot of butter in it."}}

\textit{Does \textbf{It} in "It has a lot" refer to \textbf{cake}?}
\end{quote}

\medskip
Choose \underline{No} if the pronoun does not refer to the selected word or phrase. For example,

\begin{quote}
\textit{\textcolor{mygreen}{"The large ball crashed right through the table because it was made of styrofoam."}}

\textit{Does \textbf{it} in "it was made" refer to \textbf{ball}?}
\end{quote}

\medskip
If you have any more questions, please refer to our \underline{\textcolor{blue}{\href{https://nyu-mll.github.io/SuperGLUE-human/wsc-faq}{FAQ}}} page.\medskip\\
\bottomrule
\end{tabular}
\label{tab:wsc}
\end{table}

%%%%%%%%%%%%%%%%%%%%%%%%%%%%%%%%%%%%%%%%%%%%%
%%%%%%%%%%%%%%% BoolQ %%%%%%%%%%%%%%%
%%%%%%%%%%%%%%%%%%%%%%%%%%%%%%%%%%%%%%%%%%%%%

\begin{table}[!t]
\caption{Task-specific instructions for BoolQ (continued in Table~\ref{tab:boolq1}). These instructions were provided during both training and  annotation phases.\vspace{0.2cm}}
\begin{tabular}{p{0.95\textwidth}}
\toprule
\vspace{0.1cm}
\textbf{Question-Answering Instructions}\\\\
The New York University Center for Data Science is collecting your answers for use in research on computer understanding of English. Thank you for your help!\\\\

We will present you with a passage taken from a Wikipedia article and a relevant question. \textcolor{red}{Your job is to decide, given the information provided in the passage, if the answer to the question is Yes or No.} For example, \\ \\

\textbf{In the following examples the correct answer is \underline{Yes},}
\begin{quote}
\textit{\textcolor{mygreen}{The thirteenth season of Criminal Minds was ordered on April 7, 2017, by CBS with an order of 22 episodes. The season premiered on September 27, 2017 in a new time slot at 10:00PM on Wednesday when it had previously been at 9:00PM on Wednesday since its inception. The season concluded on April 18, 2018 with a two-part season finale.}}

\textit{\textcolor{mypurple}{will there be a 13th season of criminal minds?}}

(In the above example, the first line of the passage says that the 13th season of the show was ordered.)
\end{quote}
\begin{quote}
\textit{\textcolor{mygreen}{As of 8 August 2016, the FDA extended its regulatory power to include e-cigarettes. Under this ruling the FDA will evaluate certain issues, including ingredients, product features and health risks, as well their appeal to minors and non-users. The FDA rule also bans access to minors. A photo ID is required to buy e-cigarettes, and their sale in all-ages vending machines is not permitted. The FDA in September 2016 has sent warning letters for unlawful underage sales to online retailers and retailers of e-cigarettes.}}

\textit{\textcolor{mypurple}{is vaping illegal if you are under 18?}}

(In the above example, the passage states that the "FDA rule also bans access to minors." The question uses the word "vaping," which is a synonym for e-cigrattes.)
\end{quote}

\medskip
\textbf{In the following examples the correct answer is \underline{No},}
\begin{quote}
\textit{\textcolor{mygreen}{Badgers are short-legged omnivores in the family Mustelidae, which also includes the otters, polecats, weasels, and wolverines. They belong to the caniform suborder of carnivoran mammals. The 11 species of badgers are grouped in three subfamilies: Melinae (Eurasian badgers), Mellivorinae (the honey badger or ratel), and Taxideinae (the American badger). The Asiatic stink badgers of the genus Mydaus were formerly included within Melinae (and thus Mustelidae), but recent genetic evidence indicates these are actually members of the skunk family, placing them in the taxonomic family Mephitidae.}}

\textit{\textcolor{mypurple}{is a wolverine the same as a badger?}}

(In the above example, the passage says that badgers and wolverines are in the same family, Mustelidae, which does not mean they are the same animal.)
\end{quote}

\medskip\\
% If you have any more questions, please refer to our \underline{\textcolor{blue}{\href{https://nyu-mll.github.io/SuperGLUE-human/boolq-faq}{FAQ}}} page.\medskip\\
\bottomrule
\end{tabular}
\label{tab:boolq0}
\end{table}

\begin{table}[!t]
\caption{Continuation from Table~\ref{tab:boolq0} of task-specific instructions for BoolQ. These instructions were provided during both training and  annotation phases.\vspace{0.2cm}}
\begin{tabular}{p{0.95\textwidth}}
\toprule
\vspace{0.1cm}

\begin{quote}
\textit{\textcolor{mygreen}{More famously, Harley-Davidson attempted to register as a trademark the distinctive ``chug'' of a Harley-Davidson motorcycle engine. On February 1, 1994, the company filed its application with the following description: ``The mark consists of the exhaust sound of applicant's motorcycles, produced by V-twin, common crankpin motorcycle engines when the goods are in use.'' Nine of Harley-Davidson's competitors filed oppositions against the application, arguing that cruiser-style motorcycles of various brands use the same crankpin V-twin engine which produces the same sound. After six years of litigation, with no end in sight, in early 2000, Harley-Davidson withdrew their application. }}

\textit{\textcolor{mypurple}{does harley davidson have a patent on their sound?}}

(In the above example, the passage states that Harley-Davidson applied for a patent but then withdrew, so they do not have a patent on the sound.)
\end{quote}

\medskip
If you have any more questions, please refer to our \underline{\textcolor{blue}{\href{https://nyu-mll.github.io/SuperGLUE-human/boolq-faq}{FAQ}}} page.\medskip\\
\bottomrule
\end{tabular}
\label{tab:boolq1}
\end{table}

%%%%%%%%%%%%%%%%%%%%%%%%%%%%%%%%%%%%%%%%%%%%%
%%%%%%%%%%%%%%% Diagnostic Dataset %%%%%%%%%%%%%%%
%%%%%%%%%%%%%%%%%%%%%%%%%%%%%%%%%%%%%%%%%%%%%

\begin{table}[!t]
\caption{Task-specific instructions for the diagnostic and the bias diagnostic datasets. These instructions were provided during both training and  annotation phases.\vspace{0.2cm}}
\begin{tabular}{p{0.95\textwidth}}
\toprule
\vspace{0.1cm}
\textbf{Textual Entailment Instructions}\\\\
The New York University Center for Data Science is collecting your answers for use in research on computer understanding of English. Thank you for your help!\\\\

We will present you with a prompt taken from an article someone wrote. \textcolor{red}{Your job is to figure out, based on this correct prompt (the first prompt, on top), if another prompt (the second prompt, on bottom) is also necessarily true:}\\\\ 

Choose \underline{True} if the event or situation described by the first prompt definitely implies that the second prompt, on bottom, must also be true. For example,
\begin{itemize}
    \item \textit{\textcolor{mygreen}{"Murphy recently decided to move to London."}}
    
    \textit{\textcolor{mypurple}{"Murphy recently decided to move to England."}}
    
    (The above example is True because London is in England and therefore prompt 2 is clearly implied by prompt 1.)
    
    \item \textit{\textcolor{mygreen}{"Russian cosmonaut Valery Polyakov set the record for the longest continuous amount of time spent in space, a staggering 438 days, between 1994 and 1995."}}
    
    \textit{\textcolor{mypurple}{"Russians hold record for longest stay in space."}}
    
    (The above example is True because the information in the second prompt is contained in the first prompt: Valery is Russian and she set the record for longest stay in space.)
    
    \item \textit{\textcolor{mygreen}{"She does not disgree with her brother's opinion, but she believes he's too aggresive in his defense"}}
    
    \textit{\textcolor{mypurple}{"She agrees with her brother's opinion, but she believes he's too aggresive in his defense"}}
    
    (The above example is True because the second prompt is an exact paraphrase of the first prompt, with exactly the same meaning.)
\end{itemize}

\medskip
Choose \underline{False} if the event or situation described with the first prompt on top does not necessarily imply that this second prompt must also be true. For example,
\begin{itemize}
    \item \textit{\textcolor{mygreen}{"This method was developed at Columbia and applied to data processing at CERN."}}
    
    \textit{\textcolor{mypurple}{"This method was developed at Columbia and applied to data processing at CERN with limited success."}}
    
    (The above example is False because the second prompt is introducing new information not implied in the first prompt: The first prompt does not give us any knowledge of how succesful the application of the method at CERN was.)
    
    \item \textit{\textcolor{mygreen}{"This building is very tall."}}
    
    \textit{\textcolor{mypurple}{"This is the tallest building in New York."}}
    
    (The above example is False because a building being tall does not mean it must be the tallest building, nor that it is in New York.)
    
    \item \textit{\textcolor{mygreen}{"Hours earlier, Yasser Arafat called for an end to attacks against Israeli civilians in the two weeks before Israeli elections."}}
    
    \textit{\textcolor{mypurple}{"Arafat condemned suicide bomb attacks inside Israel."}}
    
    (The above example is False because from the first prompt we only know that Arafat called for an end to attacks against Israeli citizens, we do not know what kind of attacks he may have been condemning.)
\end{itemize}

\bigskip
You do not have to worry about whether the writing style is maintained between the two prompts.\\\\

If you have any more questions, please refer to our \underline{\textcolor{blue}{\href{https://nyu-mll.github.io/SuperGLUE-human/diagnostic-faq}{FAQ}}} page.\medskip\\
\bottomrule
\end{tabular}
\label{tab:ax}
\end{table}

%%%%%%%%%%%%%%%%%%%%%%%%%%%%%%%%%%%%%%%%%%%%%
%%%%%%%%%%%%%%% GAP %%%%%%%%%%%%%%%
%%%%%%%%%%%%%%%%%%%%%%%%%%%%%%%%%%%%%%%%%%%%%

\begin{table}[!t]
\caption{Task-specific instructions for the Gendered  Ambiguous Pronoun Coreference (GAP) task. These instructions were provided during both training and  annotation phases.\vspace{0.2cm}}
\begin{tabular}{p{0.95\textwidth}}
\toprule
\vspace{0.1cm}
\textbf{GAP Instructions}\\\\
The New York University Center for Data Science is collecting your answers for use in research on computer understanding of English. Thank you for your help!\\\\

We will present you with an extract from a Wikipedia article, with one bolded pronoun. \textcolor{red}{We will also give you two names from the text that this pronoun could refer to. Your job is to figure out, based on the extract, if the pronoun refers to option A, options B, or neither:}\\\\

Choose \underline{A} if the pronoun refers to option A. For example,

\begin{quote}
\textit{\textcolor{mygreen}{"In 2010 Ella Kabambe was not the official Miss Malawi; this was Faith Chibale, but Kabambe represented the country in the Miss World pageant. At the 2012 Miss World, Susan Mtegha pushed Miss New Zealand, Collette Lochore, during the opening headshot of the pageant, claiming that Miss New Zealand was in her space."}}

\textit{Does \textbf{her} refer to option A or B below?}
\end{quote}
\begin{enumerate}[label=\Alph*]
    \item \textit{Susan Mtegha}
    \item \textit{Collette Lochore}
    \item \textit{Neither}
\end{enumerate}

\medskip
Choose \underline{B} if the pronoun refers to option B. For example,
\begin{quote}
\textit{\textcolor{mygreen}{"In 1650 he started his career as advisor in the ministerium of finances in Den Haag. After he became a minister he went back to Amsterdam, and took place as a sort of chairing mayor of this city. After the death of his brother Cornelis, De Graeff became the strong leader of the republicans. He held this position until the rampjaar."}}

\textit{Does \textbf{He} refer to option A or B below?}
\end{quote}
\begin{enumerate}[label=\Alph*]
    \item \textit{Cornelis}
    \item \textit{De Graeff}
    \item \textit{Neither}
\end{enumerate}

\medskip
Choose \underline{C} if the pronoun refers to neither option. For example,

\begin{quote}
\textit{\textcolor{mygreen}{"Reb Chaim Yaakov's wife is the sister of Rabbi Moishe Sternbuch, as is the wife of Rabbi Meshulam Dovid Soloveitchik, making the two Rabbis his uncles. Reb Asher's brother Rabbi Shlomo Arieli is the author of a critical edition of the novallae of Rabbi Akiva Eiger. Before his marriage, Rabbi Arieli studied in the Ponevezh Yeshiva headed by Rabbi Shmuel Rozovsky, and he later studied under his father-in-law in the Mirrer Yeshiva."}}

\textit{Does \textbf{his} refer to option A or B below?}
\end{quote}
\begin{enumerate}[label=\Alph*]
    \item \textit{Reb Asher}
    \item \textit{Akiva Eiger}
    \item \textit{Neither}
\end{enumerate}

\medskip
If you have any more questions, please refer to our \underline{\textcolor{blue}{\href{https://nyu-mll.github.io/SuperGLUE-human/gap-faq}{FAQ}}} page.\medskip\\
\bottomrule
\end{tabular}
\label{tab:gap}
\end{table}

% %%%%%%%%%%%%%%%%%%%%%%%%%%%%%%%%%%%%%%%%%%%%%
% %%%%%%%%%%%%%%% PAWS %%%%%%%%%%%%%%%
% %%%%%%%%%%%%%%%%%%%%%%%%%%%%%%%%%%%%%%%%%%%%%

\begin{table}[!t]
\caption{Task-specific instructions for the Paraphrase Adversaries from Word Scrambling (PAWS) task. These instructions were provided during both training and  annotation phases.\vspace{0.2cm}}
\begin{tabular}{p{0.95\textwidth}}
\toprule
\vspace{0.1cm}
\textbf{Paraphrase Detection Instructions}\\\\
The New York University Center for Data Science is collecting your answers for use in research on computer understanding of English. Thank you for your help!\\\\

We will present you with two similar sentences taken from Wikipedia articles. \textcolor{red}{Your job is to figure out if these two sentences are paraphrases of each other, and convey exactly the same meaning:}\\\\

Choose \underline{Yes} if the sentences are paraphrases and have the exact same meaning. For example,

\begin{quote}
\textit{\textcolor{mygreen}{"Hastings Ndlovu was buried with Hector Pieterson at Avalon Cemetery in Johannesburg."}}

\textit{\textcolor{mypurple}{"Hastings Ndlovu , together with Hector Pieterson , was buried at the Avalon cemetery in Johannesburg ." }}\medskip

\textit{\textcolor{mygreen}{"The complex of the Trabzon World Trade Center is close to Trabzon Airport ." }}

\textit{\textcolor{mypurple}{"The complex of World Trade Center Trabzon is situated close to Trabzon Airport ."}}
\end{quote}

\medskip
Choose \underline{No} if the two sentences are not exact paraphrases and mean different things. For example,

\begin{quote}
\textit{\textcolor{mygreen}{"She was only a few months in French service when she met some British frigates in 1809 ."}}

\textit{\textcolor{mypurple}{"She was only in British service for a few months , when in 1809 , she encountered some French frigates ."}}\medskip

\textit{\textcolor{mygreen}{"This work caused him to trigger important reflections on the practices of molecular genetics and genomics at a time when this was not considered ethical ."}}

\textit{\textcolor{mypurple}{"This work led him to trigger ethical reflections on the practices of molecular genetics and genomics at a time when this was not considered important ."}}
\end{quote}

\medskip
If you have any more questions, please refer to our \underline{\textcolor{blue}{\href{https://nyu-mll.github.io/SuperGLUE-human/paws-faq}{FAQ}}} page.\medskip\\
\bottomrule
\end{tabular}
\label{tab:paws}
\end{table}

%%%%%%%%%%%%%%%%%%%%%%%%%%%%%%%%%%%%%%%%%%%%%
%%%%%%%%%%%%%%% Quora Insincere Questions %%%%%%%%%%%%%%%
%%%%%%%%%%%%%%%%%%%%%%%%%%%%%%%%%%%%%%%%%%%%%

\begin{table}[!t]
\caption{Task-specific instructions for the Quora Insincere Questions task. These instructions were provided during both training and  annotation phases.\vspace{0.2cm}}
\begin{tabular}{p{0.95\textwidth}}
\toprule
\vspace{0.1cm}
\textbf{Insincere Questions Instructions}\\\\
The New York University Center for Data Science is collecting your answers for use in research on computer understanding of English. Thank you for your help!\\\\

We will present you with a question that someone posted on Quora. \textcolor{red}{Your job is to figure out whether or not this is a sincere question. An insincere question is defined as a question intended to make a statement rather than look for helpful answers.} Some characteristics that can signify that a question is insincere:
\begin{itemize}
    \item Has a non-neutral tone
    \begin{itemize}
        \item Has an exaggerated tone to underscore a point about a group of people
        \item Is rhetorical and meant to imply a statement about a group of people
    \end{itemize}
    \item Is disparaging or inflammatory
    \begin{itemize}
        \item Suggests a discriminatory idea against a protected class of people, or seeks confirmation of a stereotype
        \item Makes disparaging attacks/insults against a specific person or group of people
        \item Based on an outlandish premise about a group of people
        \item Disparages against a characteristic that is not fixable and not measurable
    \end{itemize}
    \item Isn't grounded in reality
    \begin{itemize}
        \item Based on false information, or contains absurd assumptions
        \item Uses sexual content (incest, bestiality, pedophilia) for shock value, and not to seek genuine answers
    \end{itemize}
\end{itemize}

\textcolor{red}{Please note that there are far fewer insincere questions than there are sincere questions! So you should expect to label most questions as sincere.}\\\\

\textbf{Examples,}\medskip\\
Choose \underline{Sincere} if you believe the person asking the question was genuinely seeking an answer from the forum. For example,

\begin{quote}
    \textit{\textcolor{mygreen}{"How do DNA and RNA compare and contrast?"}}
    
    \textit{\textcolor{mygreen}{"Are there any sports that you don't like?}}"
    
    \textit{\textcolor{mygreen}{"What is the main purpose of penance?"}}
\end{quote}

\medskip
Choose \underline{Insincere} if you believe the person asking the question was not really seeking an answer but was being inflammatory, extremely rhetorical, or absurd. For example,

\begin{quote}
\textit{\textcolor{mygreen}{"How do I sell Pakistan? I need lots of money so I decided to sell Pakistan any one wanna buy?"}}

    \textit{\textcolor{mygreen}{"If Hispanics are so proud of their countries, why do they move out?"}}
    
    \textit{\textcolor{mygreen}{"Why Chinese people are always not welcome in all countries?"}}
\end{quote}

\medskip
If you have any more questions, please refer to our \underline{\textcolor{blue}{\href{https://nyu-mll.github.io/SuperGLUE-human/quora-faq}{FAQ}}} page.\medskip\\
\bottomrule
\end{tabular}
\label{tab:quora}
\end{table}

%%%%%%%%%%%%%%%%%%%%%%%%%%%%%%%%%%%%%%%%%%%%%
%%%%%%%%%%%%%%% Ultrafine Entity Typing %%%%%%%%%%%%%%%
%%%%%%%%%%%%%%%%%%%%%%%%%%%%%%%%%%%%%%%%%%%%%

\begin{table}[!t]
\caption{Task-specific instructions for the Ultrafine Entity Typing task. These instructions were provided during both training and  annotation phases.\vspace{0.2cm}}
\begin{tabular}{p{0.95\textwidth}}
\toprule
\vspace{0.1cm}
\textbf{Entity Typing Instructions}\\\\
The New York University Center for Data Science is collecting your answers for use in research on computer understanding of English. Thank you for your help!\\\\

We will provide you with a sentence with on bolded word or phrase. We will also give you a possible tag for this bolded word or phrase. \textcolor{red}{Your job is to decide, in the context of the sentence, if this tag is correct and applicable to the bolded word or phrase:}\\\\

Choose \underline{Yes} if the tag is applicable and accurately describes the selected word or phrase. For example,

\begin{quote}
\textit{\textcolor{mygreen}{``Spain was the gold line." \textbf{It} started out with zero gold in 1937, and by 1945 it had 65.5 tons.}}

\textit{Tag: nation}
\end{quote}

\medskip
Choose \underline{No} if the tag is not applicable and does not describes the selected word or phrase. For example,

\begin{quote}
\textit{\textcolor{mygreen}{\textbf{Iraqi museum workers} are starting to assess the damage to Iraq's history.}}

\textit{Tag: organism}
\end{quote}

\medskip
If you have any more questions, please refer to our \underline{\textcolor{blue}{\href{https://nyu-mll.github.io/SuperGLUE-human/ultra-faq}{FAQ}}} page.\medskip\\
\bottomrule
\end{tabular}
\label{tab:ultra}
\end{table}

%%%%%%%%%%%%%%%%%%%%%%%%%%%%%%%%%%%%%%%%%%%%%
%%%%%%%%%%%%%%% Empathy / Distress %%%%%%%%%%%%%%%
%%%%%%%%%%%%%%%%%%%%%%%%%%%%%%%%%%%%%%%%%%%%%
\begin{table}[!t]
\caption{Task-specific instructions for the Empathetic Reaction task. These instructions were provided during both training and annotation phases.\vspace{0.2cm}}
\begin{tabular}{p{0.95\textwidth}}
\toprule
\vspace{0.1cm}
\textbf{Empathy and Distress Analysis Instructions}\\\\
The New York University Center for Data Science is collecting your answers for use in research on computer understanding of English. Thank you for your help!\\\\

We will present you with a message someone wrote after reading an article. \textcolor{red}{Your job is to figure out, based on this message, how disressed and empathetic the author was feeling. Empathy is defined as feeling warm, tender, sympathetic, moved, or compassionate. Distressed is defined as feeling worried, upset, troubled, perturbed, grieved, distrubed, or alarmed.}\\\\

\textbf{Examples,}\\
The author of the following message was not feeling empathetic at all with an \underline{empathy score of 1}, and was very distressed with a \underline{distress score of 7},

\begin{quote}
    \textit{\textcolor{mygreen}{
    "I really hate ISIS. They continue to be the stain on society by committing atrocities condemned by every nation in the world. They must be stopped at all costs and they must be destroyed so that they wont hurt another soul. These poor people who are trying to survive get killed, imprisoned, or brainwashed into joining and there seems to be no way to stop them."
    }}
\end{quote}

\medskip
The author of the following message is feeling very empathetic with an \underline{empathy score of 7} and also very distressed with a \underline{distress score of 7},

\begin{quote}
    \textit{\textcolor{mygreen}{
    "All of you know that I love birds. This article was hard for me to read because of that. Wind turbines are killing a lot of birds, including eagles. It's really very sad. It makes me feel awful. I am all for wind turbines and renewable sources of energy because of global warming and coal, but this is awful. I don't want these poor birds to die like this. Read this article and you'll see why."
    }}
\end{quote}

\medskip
The author of the following message is feeling moderately empathetic with an \underline{empathy score of 4} and moderately distressed with a \underline{distress score of 4},

\begin{quote}
    \textit{\textcolor{mygreen}{
    "I just read an article about wild fires sending a smokey haze across the state near the Appalachian mountains. Can you imagine how big the fire must be to spread so far and wide? And the people in the area obviously suffer the most. What if you have asthma or some other condition that restricts your breathing?"
    }}
\end{quote}

\medskip
The author of the following message is feeling very empathetic with an \underline{empathy score of 7} and mildly distressed with a \underline{distress score of 2},

\begin{quote}
    \textit{\textcolor{mygreen}{
    "This is a very sad article. Being of of the first female fighter pilots must have given her and her family great honor. I think that there should be more training for all pilots who deal in these acrobatic flying routines. I also think that women have just as much of a right to become a fighter pilot as men."
    }}
\end{quote}

\medskip
If you have any more questions, please refer to our \underline{\textcolor{blue}{\href{https://nyu-mll.github.io/SuperGLUE-human/empathy-faq}{FAQ}}} page.\medskip\\
\bottomrule
\end{tabular}
\label{tab:empathy}
\end{table}

\end{document}